\documentclass{article}
\usepackage[T1]{fontenc}

\usepackage{graphicx}

\usepackage{authblk}
\usepackage{url} 
\usepackage[inkscapeformat=png]{svg}
\usepackage{color}

\usepackage{fancyhdr}

\fancypagestyle{plain}{%
   \fancyhf{} 
\fancyfoot[C]{\small Preprint submitted to  the \textit{International Congress on Mathematical Software 2024} and published as a revised article in \textit{Buzzard, K., Dickenstein, A., Eick, B., Leykin, A., Ren, Y. (eds) Mathematical Software – ICMS 2024. ICMS 2024. Lecture Notes in Computer Science, vol 14749. Springer, Cham.} DOI: \url{https://doi.org/10.1007/978-3-031-64529-7_27}}

}

\date{}

\begin{document}

\title{Towards a FAIR Documentation of Workflows and Models in Applied Mathematics}

\author[1]{Marco Reidelbach\thanks{Corresponding author

}}
\author[2]{Björn Schembera}
\author[1]{Marcus Weber}

\affil[1]{\small{Mathematics of Complex Systems, Zuse Institute Berlin, 14195 Berlin, Germany \textit{\{reidelbach,weber\}@zib.de}}}
\affil[2]{\small{Institute of Applied Analysis and Numerical Simulation, University of Stuttgart, 70569 Stuttgart, Germany \textit{bjoern.schembera@mathematik.uni-stuttgart.de}}}

\maketitle              % typeset the header of the contribution
\begin{abstract}
Modeling-Simulation-Optimization workflows play a fundamental role in applied mathematics. The Mathematical Research Data Initiative, MaRDI, responded to this by developing a FAIR and machine-interpretable template for a comprehensive documentation of such workflows. MaRDMO, a Plugin for the Research Data Management Organiser, enables scientists from diverse fields to document and publish their workflows on the MaRDI Portal seamlessly using the MaRDI template. Central to these workflows are mathematical models. MaRDI addresses them with the MathModDB ontology, offering a structured formal model description. Here, we showcase the interaction between MaRDMO and the MathModDB Knowledge Graph through an algebraic modeling workflow from the Digital Humanities. This demonstration underscores the versatility of both services beyond their original numerical domain.

% 

% \keywords{Mathematical Research Data  \and MaRDMO Plugin \and MathModDB.}
\end{abstract}

\section{Introduction}
Mathematical research data holds a pivotal role in advancing scientific understanding across various disciplines, ranging from core mathematical sciences to applied fields such as Engineering, Physics, and Digital Humanities. Mathematical research data is not limited to numerical or symbolic data contained in datasets, but also to data about the model, the solution algorithm and many more appearing in the Model-Simulation-Optimization (MSO) workflow~\cite{Koprucki2018,MaRDI2022}.  In response to the intricate nature of mathematical workflows and the need for standardized documentation, the Mathematical Research Data Initiative (MaRDI~\cite{MaRDI2022}), a project under the German National Research Data Infrastructure (NFDI~\cite{Hartl2021}), aimed at developing a robust research data infrastructure for mathematics, developed a workflow documentation template adhering to the FAIR (Findable, Accessible, Interoperable, and Reusable) principles~\cite{Wilkinson2016}. This template provides a comprehensive documentation of MSO workflows in applied mathematics, including mathematical models, methods, software, hardware, input, and output data addressing specific research objectives~\cite{Boege2023}. The template is related to the developments in the NFDI4ING consortium~\cite{Schmitt2020}, where the Metadata4Ing~\cite{metadata4ing} ontology was created to capture engineering workflows. 
To facilitate the widespread adoption of this documentation template, MaRDI introduced the MaRDMO Plugin~\cite{Reidelbach2023_CoRDI}, a tool integrated in the Research Data Management Organiser (RDMO~\cite{Engelhardt2017}). Leveraging the popularity of RDMO as the most widely used software for the creation of data management plans (DMPs) in Germany~\cite{Enke2023}, MaRDMO streamlines the process of MSO workflow documentation by fetching additional information from various sources such as Wikidata~\cite{Vrandecic2012}, swMath~\cite{Greuel2014} and zbMath~\cite{Hulek2020} and making the complete documentation accessible via the MaRDI Portal\footnote{\url{https://portal.mardi4nfdi.de}}, a wikibase for Mathematics. This semi-automatic approach enhances efficiency while ensuring the completeness and accuracy of the documented workflows.

At the heart of these MSO workflows lies the proper documentation of mathematical models. Recognizing this fundamental aspect, MaRDI has developed the Mathematical Models Database (MathModDB) ontology~\cite{Schembera2023_CoRDI,Schembera2023_arxiv}, meticulously designed to capture essential elements associated with mathematical models, including research fields, problems, formulations, quantities, and tasks. Numerical models documented through this scheme will be made accessible to the public through the MathModDB Knowledge Graph (KG), ensuring widespread availability and utilization of these valuable resources.

In this paper, we aim to establish a connection between MaRDMO and the MathModDB KG, allowing researchers to access a standardized collection of mathematical models through MaRDMO. Additionally, MaRDMO can serve as an additional interface for MathModDB to gather new models from scientists across disciplines via RDMO. Through a practical example of a \textit{Logical Data Analysis} algebraic workflow from the Digital Humanities, we demonstrate how this integration facilitates comprehensive documentation and analysis, enhancing reproducibility, transparency, collaboration, and interdisciplinary innovation in the scientific community. In addition, we show the transferability of MaRDMO and MathModDB, initially developed for capturing MSO workflows and their numerical models, to algebraic modeling applications.

\section{MaRDMO, MathModDB and their Connection}

\subsection{MaRDMO} 
The MaRDMO plugin builds upon the capabilities of RDMO, which enables the creation of DMPs through customizable questionnaires. By now, general questionnaires exist reflecting the requirements of individual funding organizations, as well as subject-specific catalogs.\footnote{\url{https://github.com/rdmorganiser/rdmo-catalog}}.  With MaRDMO we introduced a questionnaire\footnote{\url{https://github.com/MarcoReidelbach/MaRDMO-Questionnaire}\label{MaRDMO_Questionnaire}} initially tailored for the documentation of MSO workflows and a customized export plugin\footnote{\url{https://github.com/MarcoReidelbach/MaRDMO-Export-Plugin}}. Such plugins extend the functionality of RDMO, e.g. facilitating an export to Zenodo\footnote{\url{https://github.com/rdmorganiser/rdmo-plugins}}. In turn, MaRDMO enables direct export to the MaRDI Portal. As far as we know, this is the first connection of RDMO with a KG. 

The MaRDMO questionnaire guides researchers through the documentation process, facilitating the capture of crucial information about their workflows. Divided into four sections, the questionnaire covers various aspects of the research process, including general aspects (1), models, variables and parameters (2), process information (3), and reproducibility considerations (4). Through a series of structured questions, researchers can input details about their research objectives, mathematical models, software and hardware used, input and output data, methods, and more (c.f. complete questionnaire\footnotemark[6]).

The guiding principle of MaRDMO is reusing existing information wherever possible from established data sources. This approach ensures proper integration into the existing research data landscape while minimizing duplication of effort. The latest version of MaRDMO has streamlined the retrieval of additional information, now seamlessly conducted in the background. This enhancement, coupled with dynamic on-the-fly queries of repositories enhances usability and functionality of the plugin. For instance, when the DOI of a publication related to the workflow is provided, MaRDMO automatically retrieves all relevant citation and author information. This information is then presented for validation and completion, ensuring proper documentation of the workflow's context.

Upon completion of the questionnaire, MaRDMO facilitates the export of the documented workflow to the MaRDI Portal. Therefore, a comprehensive summary is generated following the MaRDI template and published as wikipage. Essential details (related publication, scientific fields and mathematical areas, applied mathematical model, methods, software, input and output data) are integrated into the KG of the MaRDI Portal, making them accessible to other researchers for search and exploration. Thereby, MaRDMO empowers researchers to efficiently capture and share their MSO workflows, fostering collaboration, reproducibility, and transparency within the scientific community.

\subsection{MathModDB} 
The MathModDB ontology is designed to specifically address the intricate nature of mathematical models in the domain of MSO. Within the vast landscape of scientific reasoning, mathematical models play a fundamental role, serving as indispensable tools for abstraction, formalization, analysis, and comprehension across diverse disciplines. However, the complexity and diversity inherent in mathematical models necessitate a unified semantic framework to comprehensively capture their essence.

Recognizing this imperative, MathModDB emerges as a sophisticated ontology crafted to meet the multifaceted demands of mathematical modeling. By structuring mathematical knowledge into a coherent framework, MathModDB enhances the semantic representation of mathematical models as they appear in the MSO workflow, while fostering interoperability and accessibility across domains.

The ontology is structured around eight classes essential for describing mathematical models comprehensively, a first result of iterative (and still ongoing) development~\cite{Schembera2023_arxiv,Schembera2023_CoRDI} shaped by project internal discussions and valuable feedback from the mathematics community. These classes have evolved over time to reflect the diverse needs and perspectives within the mathematical modeling domain, ensuring a comprehensive representation of mathematical models. Moreover, this iterative process underscores the ontology's adaptability, with the potential for further evolution in response to emerging trends and evolving requirements within the mathematics community. 

The classes of MathModDB include the \textit{Mathematical Model} itself, the \textit{Research Field} within which the model operates, the \textit{Research Problem} it addresses, the \textit{Mathematical Formulations} formalizing the model, the \textit{Quantities} and \textit{Quantity Kinds} involved in the \textit{Mathematical Formulations}, the \textit{Computational Tasks} associated with the model, and the \textit{Publications} inventing, studying, surveying, or using the model. The \textit{Computational Task} class was introduced to MathModDB only recently to accommodate the diverse range of tasks or questions posed to a model, resulting in different formulations, inputs and outputs. 

\subsection{MaRDMO and MathModDB} 
The connection of MaRDMO and MathModDB marks a significant advancement in the documentation of workflows and mathematical models, so far only stemming from the domain of MSO. Prior to the integration, MaRDMO's documentation of mathematical models, not yet present in established repositories, was limited, as researchers could only provide basic information such as model name, description, main subject, defining formulas, and identifiers. This often fell short of providing a comprehensive understanding of the model, especially when considered in isolation from the contextual workflow. 

To address this limitation, the MathModDB ontology has been integrated into MaRDMO, facilitating the generation of comprehensive and insightful documentation for the models used in MSO workflows. This integration introduces additional question sets aligned with the classes of MathModDB. Following MaRDMO's guiding principle of utilizing existing resources, these question sets leverage the MathModDB KG. Now, mathematical models can be documented by connecting documented aspects of other models using the MathModDB vocabulary. In cases where no suitable entity exists, new ones can be created, with each domain featuring mandatory and optional fields. To integrate new mathematical models and their components properly into the existing data landscape, MaRDMO is also able to establish model interconnections through the MathModDB vocabulary, e.g. using generalizations, specifications or combinations. While these interconnections are crucial for maximizing the utility of the MathModDB KG, they might pose challenges for individual researchers, leaving it an optional asset. If researchers opt not to interconnect their model, the responsibility falls on domain experts to curate the MathModDB KG. This dual approach ensures that the KG remains a valuable asset while also accommodating the varying needs and perspectives of researchers.

\section{A semantic Representation of an Algebraic Modeling Workflow}

We now show a first proof of concept of how a workflow from algebraic modeling can be semantically represented by MaRDMO and MathModDB. 

\subsection{Introduction} 
The \textit{Logical Data Analysis} algebraic modeling workflow, rooted in the Digital Humanities, particularly Egyptology, intertwines the disciplines of Mathematics and Archaeology to unearth hidden patterns within a database. Originating from the Cachette de Karnak, an archaeological repository unearthed in 1903 by G. Legrain~\cite{Legrain1904}, the workflow aims to discern underlying rules governing the destruction patterns observed in ancient Egyptian objects. Legrain's findings revealed common destruction patterns such as missing heads, amputated body parts, and fragmented pieces, sparking the inquiry into whether specific rules governed these occurrences.

\subsection{Object Data and Encoding} 
An online database\footnote{\url{https://www.ifao.egnet.net/bases/cachette/}} cataloging statues discovered in the Cachette provides the foundation for the analysis. Experts from Egyptology identified 16 potentially significant properties inherent in a subset of 333 artifacts, which were then encoded into binary numbers, denoting the presence or absence of each property.

\subsection{Interdisciplinary Collaboration}
\begin{figure}
\includegraphics[width=\textwidth]{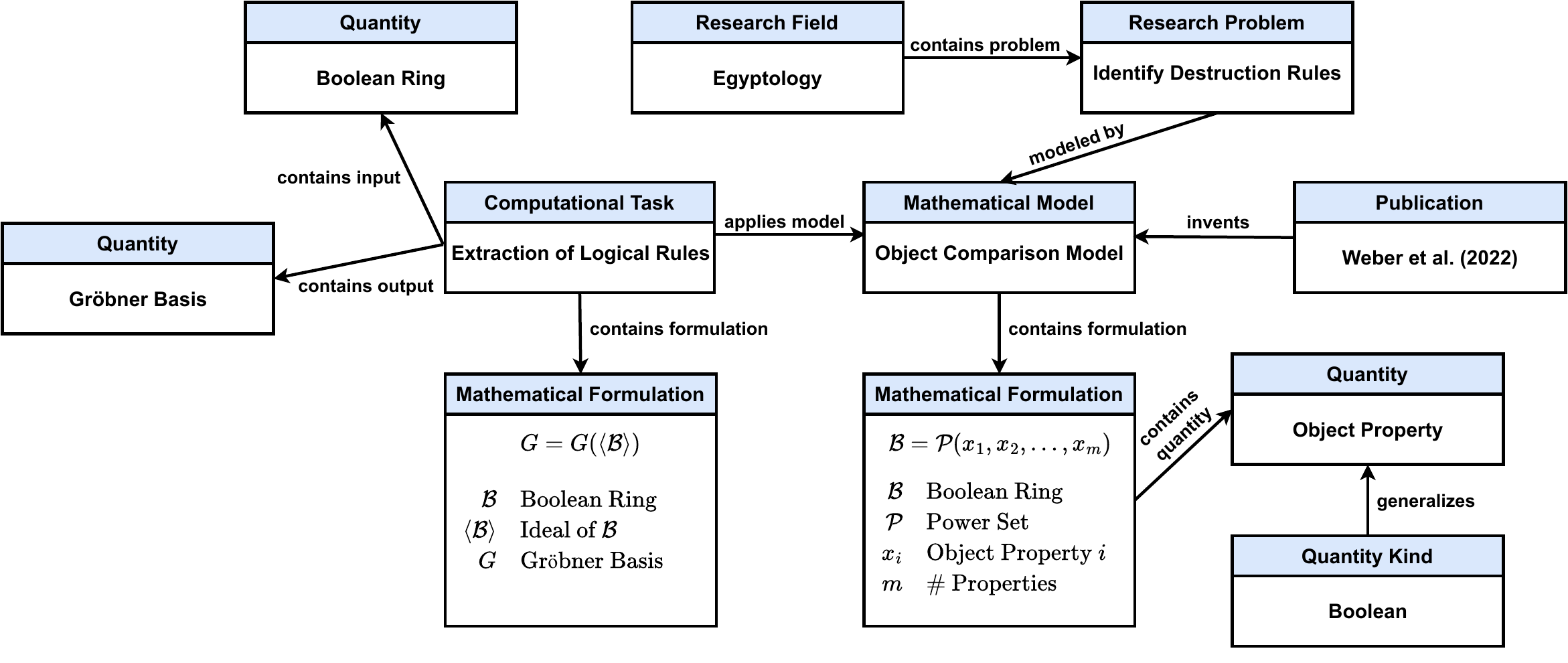}
\caption{Schematic representation of the object comparison model within the field of Egyptology, designed to identify destruction rules. Developed by Weber et al. (2022)~\cite{Weber2022}, the model employs a boolean ring as its formulation, with object properties serving as quantities, represented as booleans. The model is associated with a task, featuring its own formulation, input, and output quantities.} \label{fig1}
\end{figure}

The encoded dataset was then passed from Egyptology to Mathematics, marking the interdisciplinary exchange of data. Mathematicians then applied an object comparison model~\cite{Weber2022} (c.f. Fig \ref{fig1}), utilizing a boolean ring, to unravel the underlying rules governing the destruction patterns. Using the “Rules and Pattern” algorithm\footnote{\url{https://github.com/pynoor/The-RAP-Algorithm}}, written in Julia~\cite{Bezanson2017} and leveraging the OSCAR package~\cite{OSCAR}, the generator of the ideal was computed as a comprehensible Gröbner Basis comprising 172 logical rules.

\subsection{Interpretation and Validation}
The computational output sheds light on potential underlying rules. The validation of these expressions for scientific and statistical relevance falls upon Egyptology experts, marking the return of data to its disciplinary origin. This interdisciplinary exchange ensures a comprehensive evaluation of the identified rules.

\subsection{Reproducibility and Future Applications}
The algebraic modeling workflow presented here follows a structured process, encompassing data selection and encoding, computational analysis, and interpretation of results. The methodology is reproducible, underscoring its reliability for future studies. Beyond Archaeology, the Object Comparison Model holds promise for solving diverse computational tasks and addressing distinct problems across various domains, e.g. literary analysis~\cite{Weber2022}. Computational tasks, other than the “Extraction of Logical Rules” include object sortings and feature extractions, broadening the applicability of the model beyond its initial context.

The documentation of the \textit{Logical Data Analysis} workflow, following the MaRDI template, can be accessed through the MaRDI Portal\footnote{Documentation Video, Wikipage and Knowledge Graph Entry: \\ \url{https://portal.mardi4nfdi.de/wiki/MaRDMO} \\ \url{https://portal.mardi4nfdi.de/wiki/Logical\_Data\_Analysis\_for\_Egyptian\_Objects} \\ \url{https://portal.mardi4nfdi.de/wiki/Item:Q6032641}}. The underlying “Object Comparison Model” is visualized in Fig \ref{fig1} since the MathModDB KG is not yet publicly accessible. All aspects of the algebraic modeling workflow described before could be documented through the MaRDMO Plugin and the integrated MathModDB Ontology.

\section{Conclusion and Outlook}

The integration of MathModDB into MaRDMO significantly enhances the quality of the workflow documentation, providing a more comprehensive and insightful portrayal of applied mathematical models. Through RDMO, we expect, once again, to reach scientists from diverse disciplines and facilitate their access to standardized and enriched model documentation. 

We have shown how a research project from a seemingly completely different field, namely the humanities, can benefit from our approach. This involves the differentiation of destruction patterns in ancient objects, which can be modeled using mathematical/algebraic methods and semantically enriched and represented with the help of the tools presented (MaRDMO and MathModDB). Moreover, we have demonstrated that our solutions work beyond the domain of the classic numerical MSO workflow as they were applied to an algebraic modeling workflow. All the relevant information was captured by tools initially developed for MSO. 

Looking ahead, our next objective is to connect MaRDMO with the Mathematical Algorithms Database (MathAlgoDB~\cite{AlgoData2022}) KG\footnote{\url{https://algodata.mardi4nfdi.de/}}, also developed within MaRDI, to bolster the method component of the workflow documentation. This integration could leverage existing connections between MathAlgoDB and MathModDB, offering users a curated selection of algorithms to address specific tasks associated with particular models, thus further enhancing usability. Moreover, given the interdisciplinary nature of these workflows, we aim to collaborate with the other consortia of the NFDI to explore potential connections with non-mathematical services, expanding the scope and utility of MaRDMO in the future. In addition, further research is needed to show that the connection between MaRDMO and MathModDB also works for more advanced workflows in the field of algebra.

\subsubsection*{Acknowledgements} Marco Reidelbach, Björn Schembera and Marcus Weber are supported by MaRDI, funded by the Deutsche Forschungsgemeinschaft (DFG), project number 460135501, NFDI 29/1 “MaRDI – Mathematische Forschungsdateninitiative”.

 \bibliographystyle{splncs04}
\bibliography{FAIRWorkflowsModels}

\begin{thebibliography}{10}
\providecommand{\url}[1]{\texttt{#1}}
\providecommand{\urlprefix}{URL }
\providecommand{\doi}[1]{https://doi.org/#1}

\bibitem{metadata4ing}
Arndt, S., Farnbacher, B., Fuhrmans, M., Hachinger, S., Hickmann, J., Hoppe,
  N., et~al.: {Metadata4Ing: An Ontology for Describing the Generation of
  Research Data within a Scientific Activity.} (Feb 2022).
  \doi{10.5281/zenodo.7706017}

\bibitem{Bezanson2017}
Bezanson, J., Edelman, A., Karpinski, S., Shah, V.B.: Julia: A fresh approach
  to numerical computing. SIAM review  \textbf{59}(1),  65--98 (2017),
  \url{https://doi.org/10.1137/141000671}

\bibitem{Boege2023}
Boege, T., Fritze, R., G{\"o}rgen, C., Hanselman, J., Iglezakis, D., Kastner,
  L., et~al.: {Research-data management planning in the German mathematical
  community}. European Mathematical Society Magazine  (2023).
  \doi{10.4171/MAG/152}

\bibitem{Engelhardt2017}
Engelhardt, C., Enke, H., Klar, J., Ludwig, J., Neuroth, H.: Research data
  management organiser. In: Proceedings of the 14th International Conference on
  Digital Preservation. pp. 25--29 (2017)

\bibitem{Enke2023}
Enke, H., Hausen, D., Henzen, C., Jagusch, G., Krause, C., Sch\"onau, S.,
  et~al.: Data management planning: Concept for setting up a working group in
  the nfdi section common infrastructures. Zenodo  (2023).
  \doi{10.5281/zenodo.7540682}

\bibitem{Legrain1904}
{G. Legrain}: {Les r\'ecentes d\'ecouvertes de Karnak}. Bulletin de
  l’institut d’\'Egypte  \textbf{5},  109--120 (1904)

\bibitem{Greuel2014}
Greuel, G.M., Sperber, W.: {swMATH -- An Information Service for Mathematical
  Software}. In: {Mathematical Software -- ICMS 2014}. pp. 691--701. Springer
  Berlin Heidelberg (2014)

\bibitem{Hartl2021}
Hartl, N., W\"ossner, E., Sure-Vetter, Y.: {Nationale
  Forschungsdateninfrastruktur (NFDI)}. Informatik Spektrum  \textbf{44},
  370--373 (2021)

\bibitem{AlgoData2022}
Himpe, C., Kleikamp, H., Fritze, R., Rave, S.: {MaRDI Task Area 2 - Scientific
  Computing @ WWU Münster. AlgoData - Algorithm Knowledge Graph - Ontology
  (Version 0.1)} (2022), \url{https://mardi4nfdi.de/algodata/0.1}

\bibitem{Hulek2020}
Hulek, K., Teschke, O.: Die transformation von zbmath zu einer offenen
  plattform für die mathematik. Mitteilungen der Deutschen
  Mathematiker-Vereinigung  \textbf{28}(2),  108--111 (2020).
  \doi{doi:10.1515/dmvm-2020-0031}

\bibitem{Koprucki2018}
Koprucki, T., Kohlhase, M., Tabelow, K., M{\"u}ller, D., Rabe, F.: {Model
  Pathway Diagrams for the Representation of Mathematical Models}. Optical and
  Quantum Electronics  \textbf{50}, ~1--9 (2018).
  \doi{10.1007/s11082-018-1321-7}

\bibitem{OSCAR}
Oscar -- open source computer algebra research system, version 0.12.1-dev
  (2023), \url{https://www.oscar-system.org}

\bibitem{Reidelbach2023_CoRDI}
Reidelbach, M., Ferrer, E., Weber, M.: {MaRDMO Plugin - Document and Retrieve
  Workflows Using the MaRDI Portal}. In: {Proceedings of the 1st Conference on
  Research Data Infrastructure (CoRDI) - Connecting Communities } (2023).
  \doi{10.52825/cordi.v1i.254}

\bibitem{Schembera2023_arxiv}
Schembera, B., W{\"u}bbeling, F., Kleikamp, H., Biedinger, C., Fiedler, J.,
  Reidelbach, M., et~al.: Ontologies for models and algorithms in applied
  mathematics and related disciplines. arXiv preprint arXiv:2310.20443  (2023).
  \doi{10.48550/arXiv.2310.20443}

\bibitem{Schembera2023_CoRDI}
Schembera, B., Wübbeling, F., Koprucki, T., Biedinger, C., Reidelbach, M.,
  Schmidt, B., et~al.: {Building Ontologies and Knowledge Graphs for
  Mathematics and its Applications}. In: {Proceedings of the 1st Conference on
  Research Data Infrastructure (CoRDI) - Connecting Communities } (2023).
  \doi{10.52825/cordi.v1i.255}

\bibitem{Schmitt2020}
Schmitt, R.H., Anthofer, V., Auer, S., Ba{\c{s}}kaya, S., Bischof, C., Bronger,
  T., et~al.: {NFDI4Ing-the National Research Data Infrastructure for
  Engineering Sciences}. Zenodo  (2020). \doi{10.5281/zenodo.4015201}

\bibitem{MaRDI2022}
{The MaRDI consortium}: {MaRDI: Mathematical Research Data Initiative Proposal}
  (2022). \doi{10.5281/zenodo.6552436}

\bibitem{Vrandecic2012}
{Vrande\v{c}i\'{c}, D.}: Wikidata: a new platform for collaborative data
  collection. In: Proceedings of the 21st International Conference on World
  Wide Web. p. 1063–1064. Association for Computing Machinery (2012).
  \doi{10.1145/2187980.2188242}

\bibitem{Weber2022}
{Weber, M. and Fackeldey, K.}: {The Mathematics of Comparing Objects}. Zenodo
  (2022). \doi{doi:10.48550/arXiv.2201.07032}

\bibitem{Wilkinson2016}
Wilkinson, M.D., Dumontier, M., Aalbersberg, I.J., Appleton, G., Axton, M.,
  Baak, A., et~al.: {The FAIR Guiding Principles for scientific data management
  and stewardship}. Scientific data  \textbf{3}(1), ~1--9 (2016).
  \doi{10.1038/s41597-019-0009-6.}

\end{thebibliography}

\end{document}